\documentclass{article}

\usepackage[square,sort,comma,numbers]{natbib}

\usepackage{setspace}

\usepackage{amsmath}
\usepackage{graphicx}
\usepackage{arxiv}

\usepackage[utf8]{inputenc} 
\usepackage[T1]{fontenc}    
\usepackage{hyperref}       
\usepackage{url}            
\usepackage{booktabs}       
\usepackage{amsfonts}       
\usepackage{nicefrac}       
\usepackage{microtype}      
\usepackage{lipsum}

\usepackage{lineno,hyperref}
\usepackage{algorithm} 
\usepackage{xcolor}  
\usepackage{mathtools} 
\usepackage{amsmath,amssymb,amsfonts} 
\usepackage{algorithmic} 
\usepackage[singlelinecheck=false]{caption} 

\usepackage{adjustbox} 
\usepackage{tabularx,ragged2e,booktabs} 
\newcolumntype{L}{>{\RaggedRight\arraybackslash}X} 
\newcolumntype{C}{>{\centering\arraybackslash}X} 

\title{Topological Data Analysis in Text Classification: Extracting Features with Additive Information}
\author{
  Shafie Gholizadeh \\
  Department of Computer Science \\
  University of North Carolina at Charlotte\\
  Charlotte, NC 28223 \\
  \texttt{sgholiza@uncc.edu} \\
   \And
  Ketki Savle \\
  Department of Computer Science\\
  University of North Carolina at Charlotte\\
  Charlotte, NC 28223 \\
  \texttt{ketki.savle@gmail.com} \\
   \And
  Armin Seyeditabari \\
  Department of Computer Science\\
  University of North Carolina at Charlotte\\
  Charlotte, NC 28223 \\
  \texttt{sseyedi1@uncc.edu} \\
   \And
  Wlodek Zadrozny \\
  Department of Computer Science\\
  University of North Carolina at Charlotte\\
  Charlotte, NC 28223 \\
  \texttt{wzadrozn@uncc.edu} \\
}

\begin{document}
\maketitle

\onehalfspacing

\begin{abstract}
While the strength of Topological Data Analysis has been explored in many studies on high dimensional numeric data, it is still a challenging task to apply it to text. As the primary goal in topological data analysis is to define and quantify the shapes in numeric data, defining shapes in the text is much more challenging, even though the geometries of vector spaces and conceptual spaces are clearly relevant for information retrieval and semantics. In this paper, we examine two different methods of extraction of topological features from text, using as the underlying representations of words the two most popular methods, namely word embeddings and TF-IDF vectors. To extract topological features from the word embedding space, we interpret the embedding of a text document as high dimensional time series, and we analyze the topology of the underlying graph where the vertices correspond to different embedding dimensions. For topological data analysis with the TF-IDF representations, we analyze the topology of the graph whose vertices come from the TF-IDF vectors of different blocks in the textual document. In both cases, we apply homological persistence to reveal the geometric structures under different distance resolutions. Our results show that these topological features carry some exclusive information that is not captured by conventional text mining methods. In our experiments we observe adding topological features to the conventional features in ensemble models improves the classification results (up to 5\%). On the other hand, as expected, topological features by themselves may be not sufficient for effective classification. It is an open problem to see whether TDA features from word embeddings might be sufficient, as they seem to perform within a range of few points from top results obtained with a linear support vector classifier.
\end{abstract}


\linespread{1.5}

\section{Introduction}
\label{sec:intro}

The primary goal in Topological Data Analysis (TDA) is to analyze the shapes in data. While TDA has got significant attention in data mining for numeric data \cite{de2006coordinate, de2007coverage,  khasawneh2016chatter, khasawneh2018topological, pereira2015persistent, maletic2016persistent}, its application in natural language processing still appears to be challenging. Defining shapes in the text seems much more challenging even though vector spaces are used as standard tools to define geometries in text mining and information retrieval \cite{manning2008introduction}, and conceptual spaces \cite{gardenfors2014geometry} are relevant for cognitive modeling and semantics of natural language.

In this study, the focus is on introducing and examining two methods to apply TDA to text classification. Term frequency (or TF-IDF) and word embeddings are the most frequently used methods to translate the text into numerical data. Therefore they deserve to be examined, as a priority, for potential to reveal their hidden dimensions by applying topological methods. 

First, we introduce a novel method of using word embeddings where we view text documents as time series. We believe this method shows great promise, since it can be applied to documents irrespective of their length (with some likely limitations explained in Section \ref{sec:conclusion}), and it encodes the temporal succession words in a latent semantic space. Our algorithm analyzes the topology of the embedding space to discover relations among different embedding dimensions of the analyzed text. The precise nature of this space is not clear to us at this point. However, we know it is there, because our experiments show its influence on the accuracy of classification. 

In the second experiment, working with TF-IDF representations of textual documents, we use a  method that divides the text to a fixed number of blocks, analyzes the topological structure of the relations among different blocks and summarizes the results. As with the first method, this topological summary consists of numerical features derived from the persistence diagram of each document. And as in the first case, it improves the accuracy of classification, proving the existence of the latent topological dimension (speaking metaphorically). 

The intuitive idea behind both experiments has to do with the central premise of topological data analysis, namely that when examining a cloud of data points at different resolutions, the emerging diagrams encode global geometric properties of the point as shown in Fugure  \ref{fig:diagram} and later in Fugure  \ref{fig:ph}. There we observe, with the change of the threshold, i.e. the distance at which we add connections to the points, new elements are added to the persistence diagram, culminating in a clear circle, or torus-like  signature in in Fugure  \ref{fig:diagram} (shown as the long line in the right panel), and a more complex representation of the geometry in Fugure  \ref{fig:ph}. In our case we measure, and use as features for machine learning the birth and death diameters in dimension 0 and 1, as well as their derivatives that is the number of holes, the average divided by the standard deviation of death diameter,  and the same ratio for the duration (death - birth). 

This paper is structured as follows. In Section \ref{sec:background}, we review the basic methods in topological data analysis and review a few studies that utilized TDA in the wider area of natural language processing. Section \ref{sec:method} describes our methods of extracting topological features out of textual documents. Then our experiments are explained in Section \ref{sec:data} followed by results in Section \ref{sec:results}. We discuss the contribution and its limitations and open problems in Section \ref{sec:conclusion}.

\section{Background}
\label{sec:background}

\subsection{A Brief Sketch of Topological Data Analysis (TDA)}

In TDA, we often use \emph{simplicial complexes} to study the shapes. A \emph{simplex} can be a single point (0-simplex), two connected data points (1-simplex), tree fully-connected points (2-simplex), or generally $k+1$ fully-connected data points ($k$-simplex). For consistency of the definition, we use $(-1)$-simplex to describe an empty set \cite{zomorodian2010computational}. Then we can define a simplicial complex as the union of simplices satisfying one condition: If a simplex is in the simplicial complex, then any of its subsets should also be in the complex. 

The topological characteristics of a simplicial complex can be summarized in \emph{Betti numbers} \cite{edelsbrunner2000topological}. The $i$-th Betti number is defined as the number of $i$-dimensional holes in a simplicial complex. More specifically, $\beta_0$ is the number of the connected components, $\beta_1$ is the number of $1$-$D$ holes and $\beta_2$ is the number of $2$-$D$ voids, etc. Note that For a $m$-dimensional shape, for any $n \geq m$, n-th Betti number is zero. Betti numbers for some topological shapes are shown in Table \ref{tab:betti}.

\begin{table}[ht]
    \caption{Betti numbers for some topological shapes.}
\label{tab:betti}
    \centering
    \begin{tabularx}{\textwidth}{LLCCCC}
    \hline
Order     & Type & A Point & Circle & Sphere  &  Torus \\ \hline
$\beta_0$ &  components  & 1 & 1 & 1 & 1 \\
$\beta_1$ &  loops  & 0 & 1 & 0 & 2 \\
$\beta_2$ &  voids  & 0 & 0 & 1 & 1 \\
$\beta_3$ &  3D holes  & 0 & 0 & 0 & 0 \\ \hline
    \end{tabularx}
    
\end{table}

\begin{figure*}[!ht]
\centering
\includegraphics[width = 1.0\textwidth]{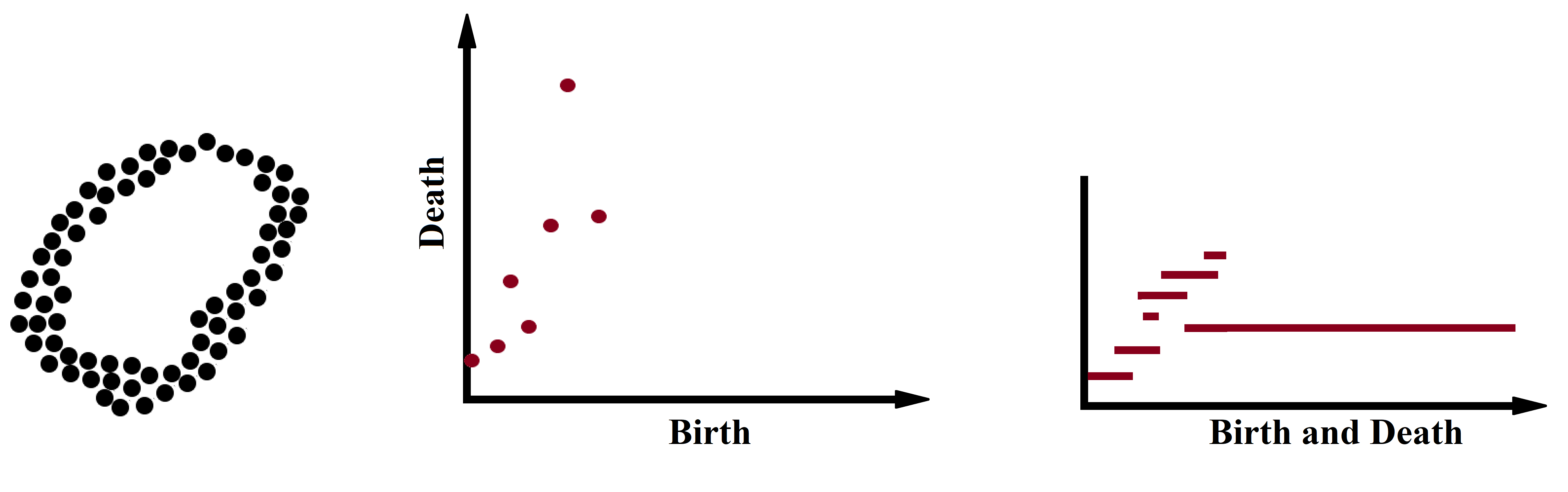}
\caption{A data cloud (left), persistence diagram (middle), and its equivalent barcodes (right). On a persistence diagram, death radii are plotted vs. birth radii. Barcodes plot the same information with one-dimensional bars from birth radii to death radii.}
\label{fig:diagram}
\end{figure*}


\emph{Persistent homology} is a technique in TDA to find topological patterns of the data  \cite{edelsbrunner2000topological,zomorodian2005computing,munch2017user}. Dealing with a set of discrete data points, we can define a radius around each data point and connect the points within that radius of each other. Then we compute the number of holes or loops in the resulting simplicial complex. However, some data points might produce a fully-connected partition where there is no hole, i.e., a $k$-simplex. If we increase the defined radius gradually, or -- equivalently -- decrease the resolution,  the resulting simplicial complex will change. Subsequently, the holes and their numbers (Betti numbers) in the shape will change. So, increasing the radius, many holes (e.g., loops in dimension $1$) will come to the picture and then will disappear. We may illustrate the birth and the death radii of the holes for each dimension in \emph{persistence diagram}  \cite{edelsbrunner2000topological}.
Equivalently, the birth and the death radii of holes might be shown with barcodes where the lifetime of every hole is represented by a one-dimensional bar from the birth radius to the death radius \cite{collins2004barcode,ghrist2008barcodes,carlsson2014topological}. An example of these barcodes and the equivalent persistence diagram are shown in Fugure  \ref{fig:diagram}.



In persistent homology, we can study the distances among data points in different ways. In the procedure we described above, the information structure based on thresholding distances is called Vietoris-Rips Filtration \cite{ghrist2008barcodes}. In a Vietoris-Rips complex, any $k$-simplex consists of $k$ nodes whose pairwise distance is less than or equal to the threshold. We refer the interested reader to \cite{zomorodian2005computing, munch2017user} for more details on persistent homology.

\subsection{TDA in Text Processing}
\label{sec:tda_text}

\begin{table*}[ht]
    \caption{Studies covering TDA in text processing.}
\label{tab:liter}
    \centering
    \begin{tabularx}{\textwidth}{llL}
    \hline
Study & Input & Task/Application \\ \hline
Wagner et al.            \cite{wagner2012computational}   & TF-IDF & Measuring heterogeneity of documents in corpus\\
Zhu                      \cite{zhu2013persistent}         & TF-IDF & Finding repetitions in text\\
Torres-Tram{\'o}n et al. \cite{torres2015topic}           & TF    & Topic detection in Twitter data\\ 
Almgren et al.           \cite{almgren2017mining}         & Word2Vec & Image popularity prediction in social media\\
Doshi and Zadrozny       \cite{doshi2018movie}            & TF-IDF & Classification\\
Gholizadeh et al.        \cite{gholizadeh2018topological} & NER  Tags   &  Authorship profiling\\
Savle and Zardozny       \cite{savle2019topological}      & TF-IDF & Text entailment prediction\\
 \hline

    \end{tabularx}
\end{table*}

There are only a few studies in the literature, utilizing topological data analysis for text processing. In most cases persistent homology is applied to term frequency vectors representing the documents. This method is used for classification \cite{doshi2018movie}, measuring heterogeneity of documents \cite{wagner2012computational}, finding repetitions in text \cite{zhu2013persistent}, topic detection \cite{torres2015topic}, and text entailment prediction \cite{savle2019topological}. In other cases, topological data analysis is applied on word embedding representations \cite{almgren2017mining} or to tagged text \cite{gholizadeh2018topological} (after performing named entity recognition). We organize these contribution in Table \ref{tab:liter}. 

In addition to the contributions mentioned in Table \ref{tab:liter}, there are several studies utilizing persistent homology in time series and system analysis 
\cite{pereira2015persistent, khasawneh2014stability, perea2015sliding, maletic2016persistent, stolz2017persistent,   garland2016exploring,aurenhammer1991voronoi, gholizadeh2018short}. These approaches are relevant for us, since texts represented by word embeddings can be viewed as time series, as we show in Section \ref{sec:method}.

\section{Methodology}
\label{sec:method}

As mentioned before, we may refine topological features out of TF-IDF space or word embedding representation of text. Here we use both approaches.

\subsection{Topological features from word embeddings}
Our method of extracting topological features from embeddings is described in Algorithm \ref{algo:emb}. Assume that a document with $T$ tokens is represented in $D$-dimensional word embedding by $\Psi_{T \times D}$. We will treat this matrix as a $D$-dimensional time series. Of course, the length of this time series is equal to $T$. Here, we intend to investigate the topological characteristics of this time-series representing the text. First, we smooth each time-series dimension (a column of $\Psi_{T \times D}$). Smoothing is a standard technique in time-series \cite{gardner2006exponential} analysis which usually reduces the noise and improves the accuracy of prediction.

To smooth each column of the embedding representation $\Psi_{T \times D}$ we can use Eq. \ref{eq:smooth}.

\begin{equation}
    \label{eq:smooth}
    \begin{split}
        \tilde{\Psi}^{(d)}_{t} &= \frac{1}{8}\Psi^{(d)}_{t-3} + \frac{1}{4}\Psi^{(d)}_{t-2} + \frac{1}{2}\Psi^{(d)}_{t-1} + \Psi^{(d)}_{t}\\ &+ \frac{1}{2}\Psi^{(d)}_{t+1} + \frac{1}{4}\Psi^{(d)}_{t+2} + \frac{1}{8}\Psi^{(d)}_{t+3}
\end{split}
\end{equation}

Then, on the smoothed matrix $\tilde{\Psi}_{T \times D}$, we calculate the distance between different embedding dimensions.
\begin{equation}
    \label{eq:dist}
\theta(\tilde{\Psi}^{(i)} , \tilde{\Psi}^{(j)}) 
       = \frac{1}{T} ||\tilde{\Psi}^{(i)}|| \cdot ||\tilde{\Psi}^{(i)}|| \{ 1 - cos(\tilde{\Psi}^{(i)} , \tilde{\Psi}^{(j)}\}
\end{equation}

Note that the measure of distance as defined in Eq. \ref{eq:dist} encodes the word order of the text. The \emph{cosine} function in the equation compares the elements on $\tilde{\Psi}^{(i)}$ and $\tilde{\Psi}^{(j)}$ with the same indices, e.g., the first element on $\tilde{\Psi}^{(i)}$ is compared with the first element on $\tilde{\Psi}^{(j)}$, etc. But recall that in the smoothing step as in Eq. \ref{eq:smooth}, each index on $\Psi^{(i)}$ is compounded with a two lags and two lead indices. Therefore, comparing the same index of the smoothed columns $i$ and $j$ is in fact comparing each index of $i$-th original (non-smoothed) column with a few indices of $j$-th original column. That is how the cosine function encoded the word order in the algorithm. The distance as defined in Eq. \ref{eq:dist} also takes into account the magnitude of the columns (dimension) that are being compared.

The pair-wise distance matrix $\Theta_{D\times{D}} = [\theta(\tilde{\Psi}^{(i)} , \tilde{\Psi}^{(j)})]$ can be interpreted as an adjacency matrix of a graph. Thus we can easily apply persistent homology on it and get persistence diagrams at dimension $0$ (components) and dimension $1$ (loops). Then for each embedding dimension, we exclude the corresponding vertex of the graph and measure the change in persistence diagrams. These measures represent the sensitivity of the graph to each embedding dimension. We know that word embeddings are representing the tokens of the text. But our main goal is to provide a new representation for the whole document. 

To this end, the document is translated into the graph with adjacency matrix $\Theta_{D\times{D}} $ in Steps 1-6 of Algorithm \ref{algo:emb}, and then into a persistence diagram. We assume that the sensitivity of the diagram with respect to each embedding dimension represents the significance of that embedding dimension in the diagram, and therefore in the original document. 

This means that, effectively, we will be classifying documents based on the significance of each embedding dimension. Since the embeddings used to represent the words are derived from a large corpus of text, and they encode similarities and differences between contexts of words in the corpus, we are employing this latent knowledge in a way similar to the standard use of TF-IDF weighting in information retrieval. In other words, $\Omega_0$ and $\Omega_1$ represent the importance of particular dimensions in a document, similarly to the TF-IDF values representing the importance of words.

We use \emph{Wasserstein distance} \cite{edelsbrunner2010computational, berwald2018computing, cohen2010lipschitz} in Algorithm \ref{algo:emb}. It measures the minimum cost to map a distribution to another one. It is also a common metric to quantify the difference among persistence diagrams \cite{marchese2017k}. Remember that the persistence diagrams are in fact a few dots on the 2D space. To compare two persistence diagrams, Wasserstein distance measures the minimum cost of moving the dots in the first diagram to convert it to the second diagram.

Finally, as shown in Algorithm \ref{algo:emb}, we get $D$ features for topological dimension $0$ (components) and another $D$ features for topological dimension $1$ (loops). We use the resulted $2 \times D$ topological features to represent the text.

\begin{algorithm}
 \caption{Topological Features from Word Embedding}
 \label{algo:emb}
 \begin{algorithmic}[1]
 \renewcommand{\algorithmicrequire}{\textbf{Input:}}
 \renewcommand{\algorithmicensure}{\textbf{Output:}}
 
 \REQUIRE word embedding representation of text:\\
 A matrix $\Psi_{T \times D}$ where $T$ is the number of tokens in the text and $D$ is the dimentionality of word embedding.  
 
 \ENSURE  embedding-based topological features of text:\\
 A vector of size $2 \times D$.\\

  \FOR {$d = 1$ to $D$}
  \STATE Smooth d-th column of $\Psi$:  Update the smoothed matrix $\tilde{\Psi}$ smoothing  column $\Psi^{(d)}$ of $\Psi_{T \times D}$.
  \ENDFOR
 
  \FOR {$i,j = 1$ to $D$}
  \STATE Calculate distance between columns $\tilde{\Psi}^{(i)}$ and $\tilde{\Psi}^{(j)}$.\\
  
    $\theta(\tilde{\Psi}^{(i)} , \tilde{\Psi}^{(j)}) 
       = \frac{1}{T} ||\tilde{\Psi}^{(i)}|| \cdot ||\tilde{\Psi}^{(i)}|| \{ 1 - cos(\tilde{\Psi}^{(i)} , \tilde{\Psi}^{(j)}\}$
 \ENDFOR

 \STATE Apply persistent homology on $\Theta_{D\times{D}} = [\theta(\tilde{\Psi}^{(i)} , \tilde{\Psi}^{(j)})]$.\\
 Get persistence diagrams $PD_0(\Theta)$ and $PD_1(\Theta)$ for components and loops respectively.
 
 \FOR {$d = 1$ to $D$}
 \STATE  Make the persistence diagrams excluding $d$-th column and row from $\Theta$, i.e., $PD_0(\Theta \setminus d)$ and $PD_1(\Theta \setminus d)$.
 \STATE Calculate Wasserstein distance of persistence diagram including and excluding dimension $d$.\\
 $\omega_0(d) = Wasserstein\{PD_0(\Theta), PD_0(\Theta \setminus d)\}$\\
 $\omega_1(d) = Wasserstein\{PD_1(\Theta), PD_1(\Theta \setminus d)\}$
 \ENDFOR
 
 \RETURN $\Omega_0$ and $\Omega_1$ as\\
 $\Omega_0 = [\omega_0(d)] \hspace{10pt};\hspace{10pt} d= 1, \dots , D$\\
 $\Omega_1 = [\omega_1(d)] \hspace{10pt};\hspace{10pt} d= 1, \dots , D$
 \end{algorithmic} 
 \end{algorithm}

\subsection{Topological features from term frequency space}
To apply persistent homology on TF-IDF space, we follow the approach in \cite{zhu2013persistent}, i.e., dividing the textual document to a fixed number of blocks and then searching for repetitive patterns in the text. Our method is described in Algorithm \ref{algo:tfidf}. 

\begin{figure*}[!ht]
\centering
\includegraphics[width = 1.0\textwidth]{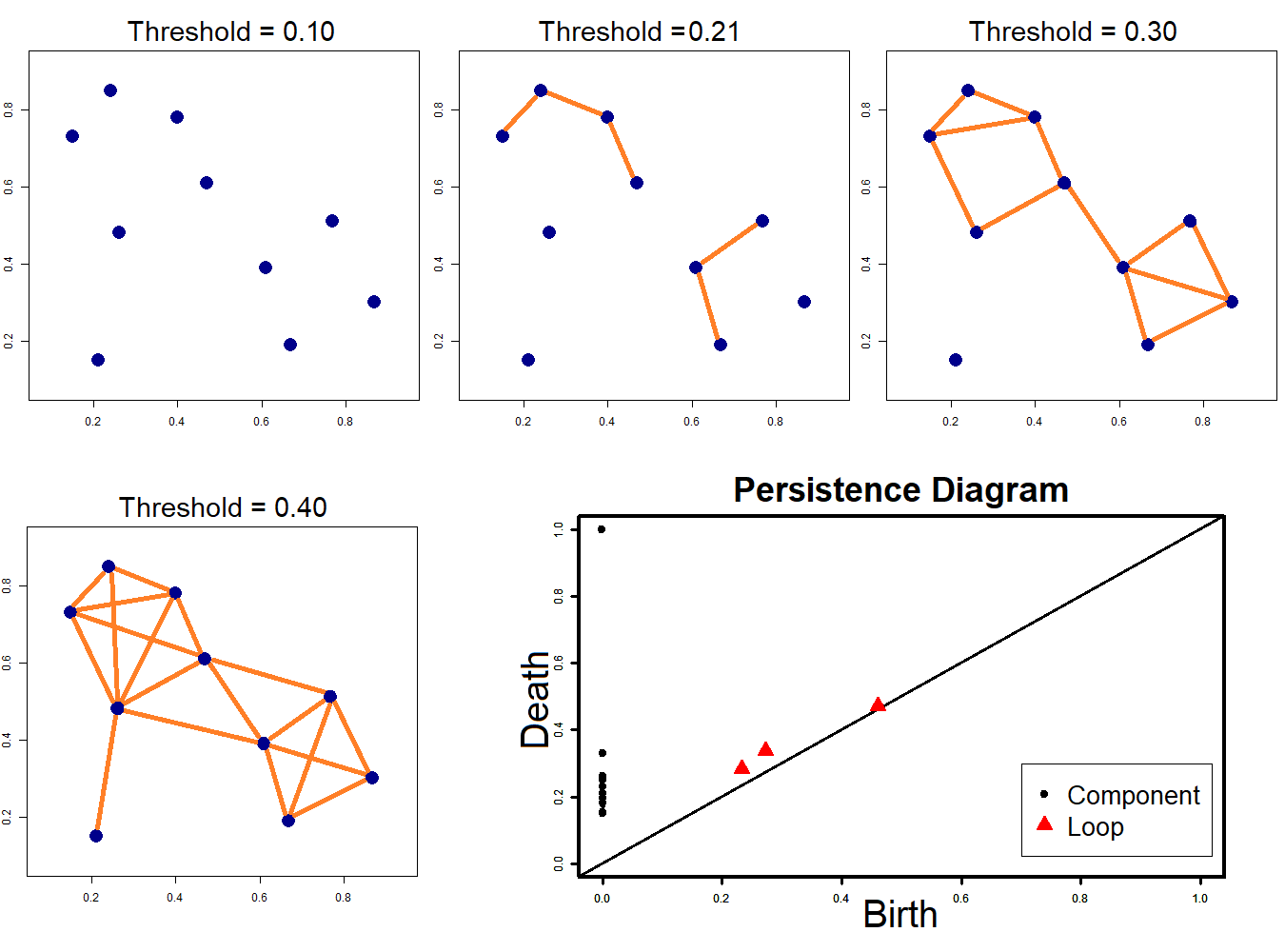}
\caption{Working on the graph of 10 vertices, persistent homology thresholds the distance (e.g., cosine distance) among different nodes using all possible thresholds. The resulted edges for a few choices of threshold are shown here. Topological characteristics are summarized in the persistence diagram (bottom right).}
\label{fig:ph}
\end{figure*}
 
We divide each document into $10$ consecutive blocks of equal size, we calculate TF-IDF vector for each block. We chose 10, but one may try different number of blocks for each document. However, we note that using a large number of blocks could make the TF-IDF vectors too sparse, so that comparing them would not be useful. For instance, if an average number of tokens in a document is only 200 tokens and we divide each of the documents into 100 blocks, there would be two tokens in each block, and most of the blocks would have zero similarity. 

In our experiments, we work on graphs of $10$ vertices, where each vertex is represented by its TF-IDF vector. An example of such graphs is illustrated in Fugure  \ref{fig:ph}. 
The figure shows that when persistent homology is applied, the number of edges connecting the ten vertices will increase with the size of the radius (as we described in Section \ref{sec:tda_text}). The distance between two vertices is given by the cosine similarity of the vectors associated with each vertex. With $10$ vertices, in topological dimension $0$ (components) we get exactly $9$ diameters of birth and $9$ diameters of death. Since for topological dimension $0$ all of the birth diameters are always equal to zero, we only retrieve $9$ death diameters. For topological dimension $1$ (loops), we may get different number of loops for different documents. Thus, if we retrieve all of birth and death diameters, we will get different numbers of features for different textual documents. Therefore, we summarize the information from topological dimension $1$ (loops) in five statistically inferred features: number of loops, the average diameter of birth, the average diameter of duration, the standard deviation of birth diameters, and the standard deviation of duration diameters.  This is similar to what Mittal and Gupta suggested in \cite{mittal2017topological} to summarize persistence diagram--- that is using six features from the persistence diagram including the number of holes, the average lifetime of holes, the maximum diameter of holes and the maximum distance between holes in each dimension. Here we utilize some similar features.  The resulting $14$ features ($9$ from dimension zero plus $5$ from dimension one) represent patterns in the text. (As noted by \cite{zhu2013persistent}  such representation may capture e.g. repetitive patterns of the text).


\begin{algorithm}
 \caption{Topological Features from TF-IDF}
 \label{algo:tfidf}
 \begin{algorithmic}[1]
 \renewcommand{\algorithmicrequire}{\textbf{Input:}}
 \renewcommand{\algorithmicensure}{\textbf{Output:}}
 
 \REQUIRE text:\\
 A array R of size T where T is number of tokens in text.  
 
 \ENSURE TF-IDF based topological features of text:\\ 
 A vector of size $14$.\\
  
  \STATE Divide R to 10 equal-size arrays of size T/10.\\
   $R = Concat(R^{(1)}, \dots, R^{(10)})$
 
  \FOR {$d = i$ to $10$}
  \STATE Calculate TF-IDF vector of $R^{(i)}$.\\
  $\phi(i) = \text{TF-IDF}\{R^{(i)}\}$
  \ENDFOR
 
  \STATE Apply persistent homology on $\Phi = [\phi(i)]$ ; $i=1, \dots , 10$.
  
  \STATE Set $X = [x_1, \dots , x_9]$ where $x_i$'s are the diameters of deaths for components (dimension $0$). We get exactly $9$ death diameters.
  
  \STATE For each loop (dimension $1$), we have the diameters of birth and death.\\
  Calculate $Y = [y_1, \dots , y_5]$\\
\begin{itemize}
    \item[] $y_1 =$  number of loops \\
    \item[] $y_2 =$ average diameter of birth \\
    \item[] $y_3 =$ average diameter of duration (death minus birth diameter) \\
    \item[] $y_4 =$ standard deviation of of birth diameters \\
    \item[] $y_5 =$ standard deviation of duration diameters
\end{itemize}

 \RETURN $X$ and $Y$
 \end{algorithmic} 
 \end{algorithm}

\section{Description of the Experiments}
\label{sec:data}

We run both algorithms on Wikipedia Movie Plots from Kaggle\footnote{https://www.kaggle.com/aminejallouli/genre-classification-based-on-wiki-movies-plots/data}. We selected the movie plots annotated by four major genres of Drama, Comedy, Action, and Romance. Keeping only the plots containing at least 200 words, we tried to predict the genres solely based on the plot texts.

The data set contains 11,500 total records. Each record may have been annotated by more than one label. More specifications per class are shown in Table \ref{tab:data}.  We used 2/3 of the records for training and 1/3 for testing. 

To represent the data in word embedding space, we used fastText \cite{bojanowski2016enriching, joulin2016bag} pre-trained on Wikipedia 2017 with the vocabulary size of $1$M and $300$d vectors\footnote{https://dl.fbaipublicfiles.com/fasttext/vectors-wiki/wiki.en.vec}. We chose fastText since in our initial experiment it showed slightly better performance compared to Google word2vec \cite{mikolov2013efficient, mikolov2013distributed, mikolov2013linguistic}, GloVe \cite{pennington2014glove}, and Conceptnet numberbatch \cite{speer2017conceptnet} pre-trained vectors. To apply persistent homology and extract topological features we utilized Ripser \cite{bauer2019ripser} package. The TF-IDF vectors for Algorithm \ref{algo:tfidf} were extracted with text2vec package \cite{selivanov2016text2vec}.

\begin{table}[ht]
    \caption{Number of records per class and overlaps among different classes.}
\label{tab:data}
    \centering
    \begin{tabularx}{\textwidth}{l C C C C }
    \hline
Specification  &  Drama  &  Comedy  &  Action  &  Romance \\ \hline
Overlap with drama  &  -  &  524  &  223  &  379 \\
Overlap with comedy  &  524  &  -  &  207  &  544 \\
Overlap with action  &  223  &  207  &  -  &  117 \\
Overlap with romance  &  379  &  544  &  117  &  - \\
Exclusive Records  &  4592  &  3302  &  1181  &  672 \\ \hline
Total Records  &  5615  &  4477  &  1658  &  1614 \\ \hline
    \end{tabularx}
\end{table}

\section{Results and Discussion}
\label{sec:results}

For each record in the data set, we computed two sets of topological features based on word embeddings as in Algorithm \ref{algo:emb} and TF-IDF space as in Algorithm \ref{algo:tfidf}. We will call these two sets of features \emph{TP1} and \emph{TP2}, respectively. 

First, we fed $TP1$ to the XGBoost \cite{chen2016xgboost} classifier with  ${max\_depth} = 2$, $eta = 1$, and $25$ iterations. Then we tried adding $TP2$ features to the same classifier to boost the results. We also tried a Bidirectional LSTM to classify the records without using our topological features. Our bidirectional LSTM model containing $64$-dimensional main layer output was trained with a batch size of $32$ in five epochs with adam optimizer \cite{kingma2014adam}. 

While bidirectional LSTM showed stronger performance than the XGBoost model feeding our topological features, we assumed that there might be some exclusive information carried by our topological features that are not captured by the LSTM. Thus we tried combining the LSTM results with the XGBoost models. As one of the easiest ways to combine the results, we fed the probabilities (not the rounded predictions) returned by the two models (LSTM and XGBoost using $TP1$ and $TP2$) to a logistic regression model. 

As shown in Table \ref{tab:classifiers}, our best ensemble model outperforms the LSTM accuracy and F1-score by 1.6\% and 5.1\%, respectively. The previous results\footnote{https://www.kaggle.com/aminejallouli/genre-classification-based-on-wiki-movies-plots/notebook} using linear Support Vector Classifier (SVC) and multinomial Na\"ive Bayes are also provided in the table. The detailed results per class are also provided in Table \ref{tab:by_class}.

Note that the topological features that we extracted from the word embedding space (i.e., $TP1$) can classify the records alone with an accuracy comparable but not equal to the LSTM. On the other hand, the topological features extracted from TF-IDF space are primarily used to reflect some repetitive patterns in the text, as Zhu \cite{zhu2013persistent} suggested in a similar study. However, as shown in Table \ref{tab:classifiers} and Table \ref{tab:by_class}, using the topological feature sets can boost the accuracy of classification in the ensemble model.

\begin{table}[ht]
    \caption{Macro-average results by different methods. The ensemble model using both embeddings and topological features improves the F1 measure by about  5\%.}
\label{tab:classifiers}
    \centering
    \begin{tabularx}{\textwidth}{ll C C C C}
    \hline
  & Classifier                      & Pre. & Rec. & F1    & Acc.\\ \hline
1 & BiLSTM                          & 68.0 & 59.7 & 0.608 & 76.2\\ 
2 & XGBoost on TP1                  & 59.6 & 53.2 & 0.560 & 71.1\\
3 & XGBoost on TP1 \& TP2          & 59.9 & 53.7 & 0.564 & 71.4\\
4 & BiLSTM + XGBoost on TP1         & 67.8 & 64.8 & 0.656 & 77.3\\
5 & BiLSTM + XGBoost on TP1 \& TP2 & 68.5 & 64.6 & \textbf{0.659} & \textbf{77.8}\\ 
  & Previous Results (Linear SVC)   &   &   &       & 73.5\\
  & Previous Results (Na\"ive Bayes)   &   &   &       & 73.3\\ \hline
    \end{tabularx}
\end{table}

\begin{table}[ht]
    \caption{Accuracy per class using different methods. Here BiLSTM, XGB, XGB2, TL4, and TL5 are the same as models 1 to 5 in Table \ref{tab:classifiers}. prev. SVC and prev. NB refer to the previous results using linear SVC and multinomial Na\"ive Bayes, respectively. We see across the board superior performance of the ensemble models with topological features.}
 
\label{tab:by_class}
    \centering
    \begin{tabularx}{\textwidth}{l  C C C C C  C C}
    \hline
Class    & BiLSTM & XGB  & XGB2 & TL4   & TL5  & prev. SVC & prev. NB \\ \hline
action   & 87.7 & 86.7 & 86.9 & \textbf{89.3} & 88.9 & 81.5 & 82.7 \\
comedy   & 75.6 & 69.0 & 69.1 & 76.9 & \textbf{77.7} & 74.6 & 73.3 \\
drama    & 69.9 & 63.9 & 64.3 & 71.0 & \textbf{71.6} & 66.1 & 67.4 \\
romance  & 87.6 & 86.0 & 85.9 & 87.8 & 87.8 & \textbf{88.3} & 84.3 \\ \hline
macro-avg & 76.2 & 71.1 & 71.4 & 77.3 & \textbf{77.8} & 73.5 & 73.3 \\ \hline
    \end{tabularx}

\end{table}

\section{Conclusions}
\label{sec:conclusion}

We first summarize our contributions and argue for the potential of topological methods to contribute to text analysis, and then discuss some limitations and open problems.

\subsection{Summary of contributions}

In this paper, we used two different methods to extract topological features from text and applied them to the task of document classification. The first method converts text, represented as a sequence of word embeddings into a high dimensional time series, which at the end is analyzed using the machinery of topological data analysis, namely homological persistence.  The second method augments the classical TF-IDF representation of the text with topological features. 

Specifically, we have leveraged existing word embeddings along with topology of text to show that such structure can carry some useful information for machine learning classifiers to learn from. To extract topological features from the word embedding space, using the high dimensional time series derived from embeddings, we measured and analyzed the topology of the graph whose vertices are different embedding dimensions. 

For topological data analysis of TF-IDF space, we analyzed the topology of the graph whose vertices are the TF-IDF vectors of different blocks of a textual document. As we have shown in the results, while a classifier utilizing only topological features may fail to outperform more conventional models like bidirectional LSTMs, these topological features are capable of carrying some exclusive information that is not captured by conventional text mining methods. Therefore, adding these features to more conventional features models can boost the results. In our experiment, adding using topological features in the ensemble model resulted in 4.9\% increase in recall, 0.5\% increase in precision, and 5.1\% increase in F1 score

Briefly, our contributions are as follows: 
\begin{itemize}
    \item We introduced a new algorithm of extracting topological features from text, namely by converting a sequence of word embeddings into a time series, and analyzing the dimensions of the resulting series for topological persistence. 
    \item This algorithm works with documents of any length and, importantly, preserves the word order in its representation. 
    \item We have shown that this new method produces features of value for the task of document classification.
    \item We showed that even if the representation of documents is derived from the standard TF/IDF matrix, similarly produced topological features improve the accuracy of classification. 
\end{itemize}
Based on the above, we suggest that topological methods deserve deeper examination as a tool for text analysis. We believe that as with the geometries of vector spaces and conceptual spaces mentioned earlier in Section \ref{sec:intro}, the topological features, which capture certain geometric invariants are relevant for text analytics and semantics of natural language.

\subsection{Discussion, Limitations and Open Problems}

We end with a discussion, including some of the limitations, and open problems.

The strength of our algorithm for analyzing documents as a time series of embeddings is in its universal applicability, irrespective of the length of the document. The second important property is using the word order. Finally, the algorithm produces the representation in one pass. 

However, one of the limitations of our methodology is the size of block of text. Regarding the embedding based topological features, the topological structure of a short text would not be stable. Also due to lack of context, the embedding may not be able to provide enough information  for classification tasks.

Similarly, using  its TF-IDF vectors on short documents, can result in poor simplicial shapes, when we divide our text in blocks of 10, as in Section \ref{sec:method}. That is, a set of separate dots in the space most of which are not connected at all. In such a case, it is challenging to find informative topological structure in text. 

Proving the value of the methods used in this article for other natural language processing tasks, such as summarization, entity extraction or question answering, is both a limitation of this work, and an open problem.

We see two other important open issues, one very technical, and one more programmatic. The latter one has to do with connecting our work on topology of text with the work on the understanding of topological properties of deep neural networks, exemplified e.g. by \cite{kim2020efficient} and \cite{guss2018characterizing}. 
An urgent technical open problem is to find the actual text behind the topological structures. This a challenge in our ongoing work.

\bibliographystyle{unsrt}
\bibliography{main}

\begin{thebibliography}{10}

\bibitem{de2006coordinate}
Vin De~Silva and Robert Ghrist.
\newblock Coordinate-free coverage in sensor networks with controlled
  boundaries via homology.
\newblock {\em The International Journal of Robotics Research},
  25(12):1205--1222, 2006.

\bibitem{de2007coverage}
Vin De~Silva, Robert Ghrist, et~al.
\newblock Coverage in sensor networks via persistent homology.
\newblock {\em Algebraic \& Geometric Topology}, 7(1):339--358, 2007.

\bibitem{khasawneh2016chatter}
Firas~A Khasawneh and Elizabeth Munch.
\newblock Chatter detection in turning using persistent homology.
\newblock {\em Mechanical Systems and Signal Processing}, 70:527--541, 2016.

\bibitem{khasawneh2018topological}
Firas~A Khasawneh and Elizabeth Munch.
\newblock Topological data analysis for true step detection in periodic
  piecewise constant signals.
\newblock {\em Proceedings of the Royal Society A: Mathematical, Physical and
  Engineering Sciences}, 474(2218):20180027, 2018.

\bibitem{pereira2015persistent}
C{\'a}ssio~MM Pereira and Rodrigo~F de~Mello.
\newblock Persistent homology for time series and spatial data clustering.
\newblock {\em Expert Systems with Applications}, 42(15-16):6026--6038, 2015.

\bibitem{maletic2016persistent}
Slobodan Maleti{\'c}, Yi~Zhao, and Milan Rajkovi{\'c}.
\newblock Persistent topological features of dynamical systems.
\newblock {\em Chaos: An Interdisciplinary Journal of Nonlinear Science},
  26(5):053105, 2016.

\bibitem{manning2008introduction}
Christopher~D Manning, Prabhakar Raghavan, and Hinrich Sch{\"u}tze.
\newblock {\em Introduction to information retrieval}.
\newblock Cambridge university press, 2008.

\bibitem{gardenfors2014geometry}
Peter G{\"a}rdenfors.
\newblock The geometry of meaning.
\newblock {\em Semantics Based on Conceptual Spaces}, 2014.

\bibitem{zomorodian2010computational}
Afra Zomorodian.
\newblock Computational topology.
\newblock In {\em Algorithms and theory of computation handbook}, pages
  3.3--3.4. Chapman \& Hall/CRC, 2010.

\bibitem{edelsbrunner2000topological}
Herbert Edelsbrunner, David Letscher, and Afra Zomorodian.
\newblock Topological persistence and simplification.
\newblock In {\em Foundations of Computer Science, 2000. Proceedings. 41st
  Annual Symposium on}, pages 454--463. IEEE, 2000.

\bibitem{zomorodian2005computing}
Afra Zomorodian and Gunnar Carlsson.
\newblock Computing persistent homology.
\newblock {\em Discrete \& Computational Geometry}, 33(2):249--274, 2005.

\bibitem{munch2017user}
Elizabeth Munch.
\newblock A user’s guide to topological data analysis.
\newblock {\em Journal of Learning Analytics}, 4(2):47--61, 2017.

\bibitem{collins2004barcode}
Anne Collins, Afra Zomorodian, Gunnar Carlsson, and Leonidas~J Guibas.
\newblock A barcode shape descriptor for curve point cloud data.
\newblock {\em Computers \& Graphics}, 28(6):881--894, 2004.

\bibitem{ghrist2008barcodes}
Robert Ghrist.
\newblock Barcodes: the persistent topology of data.
\newblock {\em Bulletin of the American Mathematical Society}, 45(1):61--75,
  2008.

\bibitem{carlsson2014topological}
Gunnar Carlsson.
\newblock Topological pattern recognition for point cloud data.
\newblock {\em Acta Numerica}, 23:289--368, 2014.

\bibitem{wagner2012computational}
Hubert Wagner, Pawe{\l} D{\l}otko, and Marian Mrozek.
\newblock Computational topology in text mining.
\newblock In {\em Computational Topology in Image Context}, pages 68--78.
  Springer, 2012.

\bibitem{zhu2013persistent}
Xiaojin Zhu.
\newblock Persistent homology: An introduction and a new text representation
  for natural language processing.
\newblock In {\em IJCAI}, pages 1953--1959, 2013.

\bibitem{torres2015topic}
Pablo Torres-Tram{\'o}n, Hugo Hromic, and Bahareh~Rahmanzadeh Heravi.
\newblock Topic detection in twitter using topology data analysis.
\newblock In {\em International Conference on Web Engineering}, pages 186--197.
  Springer, 2015.

\bibitem{almgren2017mining}
Khaled Almgren, Minkyu Kim, and Jeongkyu Lee.
\newblock Mining social media data using topological data analysis.
\newblock In {\em Information Reuse and Integration (IRI), 2017 IEEE
  International Conference on}, pages 144--153. IEEE, 2017.

\bibitem{doshi2018movie}
Pratik Doshi and Wlodek Zadrozny.
\newblock Movie genre detection using topological data analysis.
\newblock In {\em International Conference on Statistical Language and Speech
  Processing}, pages 117--128. Springer, 2018.

\bibitem{gholizadeh2018topological}
Shafie Gholizadeh, Armin Seyeditabari, and Wlodek Zadrozny.
\newblock Topological signature of 19th century novelists: Persistent homology
  in text mining.
\newblock {\em Big Data and Cognitive Computing}, 2(4):33, 2018.

\bibitem{savle2019topological}
Ketki Savle, Wlodek Zadrozny, and Minwoo Lee.
\newblock Topological data analysis for discourse semantics?
\newblock In {\em Proceedings of the 13th International Conference on
  Computational Semantics-Student Papers}, pages 34--43, 2019.

\bibitem{khasawneh2014stability}
Firas~A Khasawneh and Elizabeth Munch.
\newblock Stability determination in turning using persistent homology and time
  series analysis.
\newblock In {\em ASME 2014 International Mechanical Engineering Congress and
  Exposition}, pages V04BT04A038--V04BT04A038. American Society of Mechanical
  Engineers, 2014.

\bibitem{perea2015sliding}
Jose~A Perea and John Harer.
\newblock Sliding windows and persistence: An application of topological
  methods to signal analysis.
\newblock {\em Foundations of Computational Mathematics}, 15(3):799--838, 2015.

\bibitem{stolz2017persistent}
Bernadette~J Stolz, Heather~A Harrington, and Mason~A Porter.
\newblock Persistent homology of time-dependent functional networks constructed
  from coupled time series.
\newblock {\em Chaos: An Interdisciplinary Journal of Nonlinear Science},
  27(4):047410, 2017.

\bibitem{garland2016exploring}
Joshua Garland, Elizabeth Bradley, and James~D Meiss.
\newblock Exploring the topology of dynamical reconstructions.
\newblock {\em Physica D: Nonlinear Phenomena}, 334:49--59, 2016.

\bibitem{aurenhammer1991voronoi}
Franz Aurenhammer.
\newblock Voronoi diagrams--a survey of a fundamental geometric data structure.
\newblock {\em ACM Computing Surveys (CSUR)}, 23(3):345--405, 1991.

\bibitem{gholizadeh2018short}
Shafie Gholizadeh and Wlodek Zadrozny.
\newblock A short survey of topological data analysis in time series and
  systems analysis.
\newblock {\em arXiv preprint arXiv:1809.10745}, 2018.

\bibitem{gardner2006exponential}
Everette~S Gardner~Jr.
\newblock Exponential smoothing: The state of the art—part ii.
\newblock {\em International journal of forecasting}, 22(4):637--666, 2006.

\bibitem{edelsbrunner2010computational}
Herbert Edelsbrunner and John Harer.
\newblock {\em Computational topology: an introduction}.
\newblock American Mathematical Soc., 2010.

\bibitem{berwald2018computing}
Jesse~J Berwald, Joel~M Gottlieb, and Elizabeth Munch.
\newblock Computing wasserstein distance for persistence diagrams on a quantum
  computer.
\newblock {\em arXiv preprint arXiv:1809.06433}, 2018.

\bibitem{cohen2010lipschitz}
David Cohen-Steiner, Herbert Edelsbrunner, John Harer, and Yuriy Mileyko.
\newblock Lipschitz functions have l p-stable persistence.
\newblock {\em Foundations of computational mathematics}, 10(2):127--139, 2010.

\bibitem{marchese2017k}
Andrew Marchese, Vasileios Maroulas, and Josh Mike.
\newblock K- means clustering on the space of persistence diagrams.
\newblock In {\em Wavelets and Sparsity XVII}, volume 10394, page 103940W.
  International Society for Optics and Photonics, 2017.

\bibitem{mittal2017topological}
Khushboo Mittal and Shalabh Gupta.
\newblock Topological characterization and early detection of bifurcations and
  chaos in complex systems using persistent homology.
\newblock {\em Chaos: An Interdisciplinary Journal of Nonlinear Science},
  27(5):051102, 2017.

\bibitem{bojanowski2016enriching}
Piotr Bojanowski, Edouard Grave, Armand Joulin, and Tomas Mikolov.
\newblock Enriching word vectors with subword information.
\newblock {\em arXiv preprint arXiv:1607.04606}, 2016.

\bibitem{joulin2016bag}
Armand Joulin, Edouard Grave, Piotr Bojanowski, and Tomas Mikolov.
\newblock Bag of tricks for efficient text classification.
\newblock {\em arXiv preprint arXiv:1607.01759}, 2016.

\bibitem{mikolov2013efficient}
Tomas Mikolov, Kai Chen, Greg Corrado, and Jeffrey Dean.
\newblock Efficient estimation of word representations in vector space.
\newblock {\em arXiv preprint arXiv:1301.3781}, 2013.

\bibitem{mikolov2013distributed}
Tomas Mikolov, Ilya Sutskever, Kai Chen, Greg~S Corrado, and Jeff Dean.
\newblock Distributed representations of words and phrases and their
  compositionality.
\newblock In {\em Advances in neural information processing systems}, pages
  3111--3119, 2013.

\bibitem{mikolov2013linguistic}
Tom{\'a}{\v{s}} Mikolov, Wen-tau Yih, and Geoffrey Zweig.
\newblock Linguistic regularities in continuous space word representations.
\newblock In {\em Proceedings of the 2013 conference of the north american
  chapter of the association for computational linguistics: Human language
  technologies}, pages 746--751, 2013.

\bibitem{pennington2014glove}
Jeffrey Pennington, Richard Socher, and Christopher Manning.
\newblock Glove: Global vectors for word representation.
\newblock In {\em Proceedings of the 2014 conference on empirical methods in
  natural language processing (EMNLP)}, pages 1532--1543, 2014.

\bibitem{speer2017conceptnet}
Robert Speer, Joshua Chin, and Catherine Havasi.
\newblock Conceptnet 5.5: An open multilingual graph of general knowledge.
\newblock In {\em Thirty-First AAAI Conference on Artificial Intelligence},
  2017.

\bibitem{bauer2019ripser}
Ulrich Bauer.
\newblock Ripser: efficient computation of vietoris-rips persistence barcodes.
\newblock {\em arXiv preprint arXiv:1908.02518}, 2019.

\bibitem{selivanov2016text2vec}
Dmitriy Selivanov and Qing Wang.
\newblock text2vec: Modern text mining framework for r.
\newblock {\em Computer software manual](R package version 0.4. 0). Retrieved
  from https://CRAN. R-project. org/package= text2vec}, 2016.

\bibitem{chen2016xgboost}
Tianqi Chen and Carlos Guestrin.
\newblock Xgboost: A scalable tree boosting system.
\newblock In {\em Proceedings of the 22nd acm sigkdd international conference
  on knowledge discovery and data mining}, pages 785--794. ACM, 2016.

\bibitem{kingma2014adam}
Diederik~P Kingma and Jimmy Ba.
\newblock Adam: A method for stochastic optimization.
\newblock {\em arXiv preprint arXiv:1412.6980}, 2014.

\bibitem{kim2020efficient}
Kwangho Kim, Jisu Kim, Joon~Sik Kim, Frederic Chazal, and Larry Wasserman.
\newblock Efficient topological layer based on persistent landscapes.
\newblock {\em arXiv preprint arXiv:2002.02778}, 2020.

\bibitem{guss2018characterizing}
William~H Guss and Ruslan Salakhutdinov.
\newblock On characterizing the capacity of neural networks using algebraic
  topology.
\newblock {\em arXiv preprint arXiv:1802.04443}, 2018.

\end{thebibliography}

\end{document}